\def\year{2021}\relax
\documentclass[letterpaper]{article} 
\usepackage{aaai21}  
\usepackage{times}  
\usepackage{helvet} 
\usepackage{courier}  
\usepackage[hyphens]{url}  
\usepackage{graphicx} 
\usepackage{booktabs}
\usepackage{amssymb}
\usepackage{natbib}  
\usepackage{amsmath}

\usepackage{caption} 
\title{Semi-Supervised Object Detection with Adaptive Class-Rebalancing Self-Training}

\title{Semi-Supervised Object Detection with Adaptive Class-Rebalancing Self-Training}
\author {
    Fangyuan Zhang,\textsuperscript{\rm 1 \rm 2}
    Tianxiang Pan, \textsuperscript{\rm 1 \rm 3}
    Bin Wang \textsuperscript{\rm 1 \rm 4} \\
} 
\affiliations {
    \textsuperscript{\rm 1} 
    Software School of Tsinghua University\\
    \textsuperscript{\rm 2} zhangfy19@mails.tsinghua.edu.cn \\
    \textsuperscript{\rm 3} ptx9363@gmail.com \\
    \textsuperscript{\rm 4} wangbins@mails.tsinghua.edu.cn
}

\begin{document}

\maketitle
\begin{abstract}

This study delves into semi-supervised object detection (SSOD) to improve detector performance with additional unlabeled data. State-of-the-art SSOD performance has been achieved recently by self-training, in which training supervision consists of ground truths and pseudo-labels.  In current studies, we observe that class imbalance in SSOD severely impedes the effectiveness of self-training.  To address the class imbalance, we propose adaptive class-rebalancing self-training (ACRST) with a novel memory module called CropBank.  ACRST adaptively rebalances the training data with foreground instances extracted from the CropBank, thereby alleviating the class imbalance. Owing to the high complexity of detection tasks, we observe that both self-training and data-rebalancing suffer from noisy pseudo-labels in SSOD.  Therefore, we propose a novel two-stage filtering algorithm to generate accurate pseudo-labels. Our method achieves satisfactory improvements on MS-COCO and VOC benchmarks. When using only 1\% labeled data in MS-COCO, our method achieves 17.02 mAP improvement over supervised baselines, and 5.32 mAP improvement compared with state-of-the-art methods. 

\end{abstract}

\section{Introduction}

Object detection to classify and localize objects in the image is one of the most important research topics in computer vision. In recent years, significant progress has been witnessed in deep-learning-based object detection. The majority of the existing studies follow a fully supervised setting and heavily rely on large datasets with bounding-box annotations. However, creating fully annotated detection datasets costs thousands of hours~\cite{7298824,5975165}, thereby hindering the practicability of current studies. Therefore, a surge of attention has been dedicated to semi-supervised object detection (SSOD). Although SSOD has made immense progress, the current SSOD methods and their fully supervised counterparts continue to have a significant performance gap.

State-of-the-art SSOD performance has been achieved recently by the self-training paradigm, in which pseudo-labels of unlabeled data are generated to train detectors. However, the majority of advanced self-training algorithms~\cite{tarvainen2018mean,xie2020selftraining,laine2017temporal} are designed specifically for classification. In the experiments, we observe that using them directly is suboptimal when \emph{class imbalance} in SSOD considerably hinders the use of self-training. 

\begin{figure*}[h!]
    \centering
    \begin{minipage}{8cm}
      \centerline{\includegraphics[width=8.0cm]{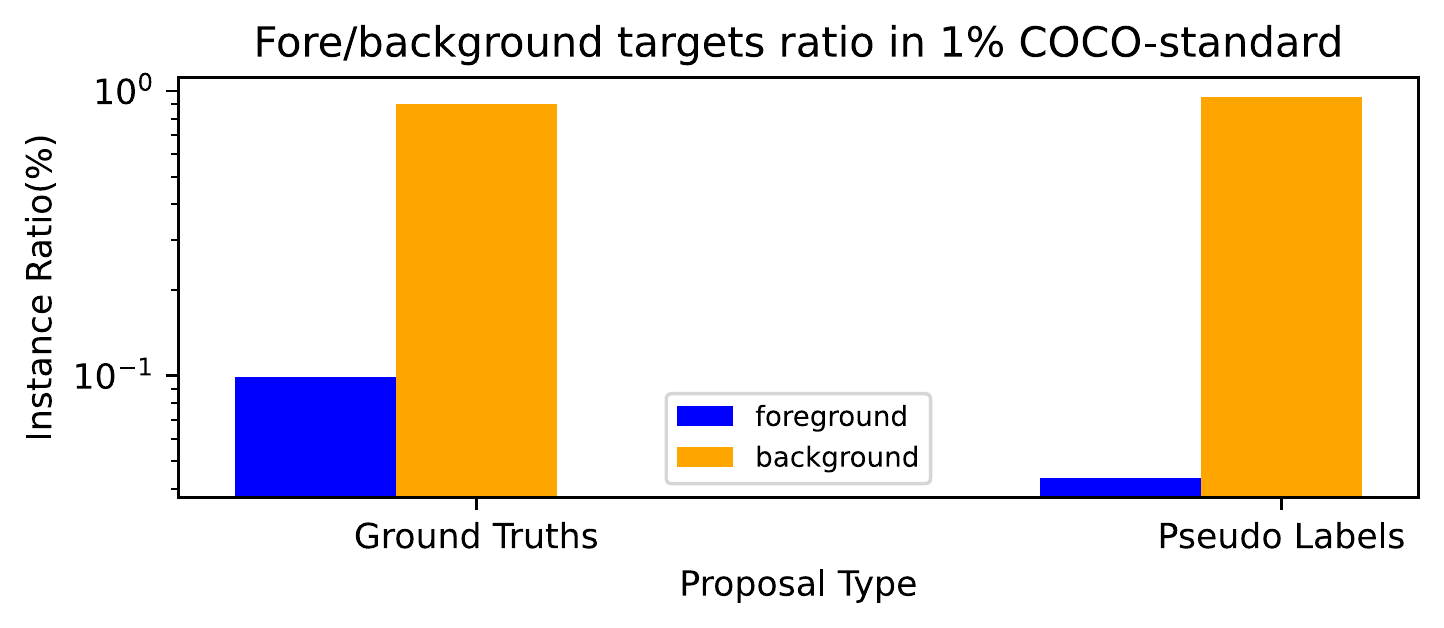}}
        \centerline{(a)}
    \end{minipage}
    \begin{minipage}{8cm}
      \centerline{\includegraphics[width=8.0cm]{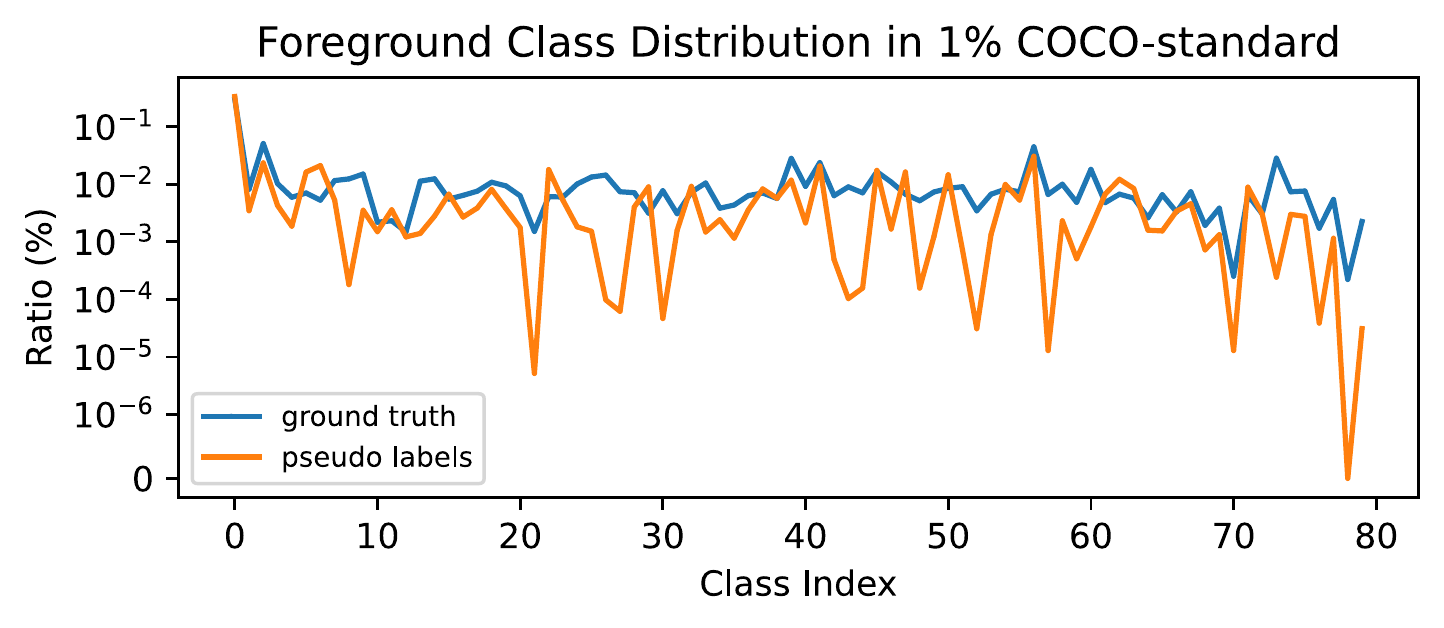}}
        \centerline{(b)}
    \end{minipage}
    
    \caption{Class imbalance in pseudo-labels in 1\% COCO-standard. (a) Foreground-background imbalance. In Faster-RCNN, background instances are predominant in the training targets of pseudo-labels. (b) Foreground-foreground imbalance is compounded in pseudo-labels.}
    \label{fig:1}
\end{figure*}

Class imbalance is a longstanding challenge in object detection that mainly includes foreground-background imbalance~\cite{2017Focal,2017Faster,chen2020accurate,cao2019prime} and foreground-foreground imbalance~\cite{peng2020largescale,imbalance}. As presented in Figure~\ref{fig:1}(a), background instances are predominant in the training targets (background instances account for $90\%$ of all training instances) when the foreground-background imbalance exists in detection data. This problem is compounded in pseudo-labels (foreground instances only account for $5\%$ of all training instances). 

Apart from foreground-background imbalance, we find a severe foreground-foreground imbalance problem in SSOD. As shown in Figure~\ref{fig:1}(b), there are some neglected classes in pseudo-labels in $1\%$ COCO-standard. The cause of this balance is two folds. First, pseudo-labels in SSOD are inaccurate due to the high complexity of the detection task. Moreover, the model trained on foreground-background imbalanced labeled data is prone to generate biased predictions.

The above two-type imbalance yields biased pseudo-labels in self-training-based SSOD. Subsequent training on biased pseudo-labels further intensifies the class imbalance, thereby aggravating the performance of the final model. 

To address the preceding issues, an intuitive idea is to utilize data-rebalancing algorithms in classification tasks ~\cite{2020Libra,2016Factors,2017Faster}. However, this idea is impeded by entanglements of foreground-background and foreground-foreground instances in detection data. To decouple these entanglements, we introduce a novel memory module called \emph{CropBank} to store ground truths/pseudo labels of foreground instances in labeled/unlabeled data. With the CropBank, we propose two detection-specific data-rebalancing algorithms: foreground-background rebalancing (FBR) and adaptive foreground-foreground rebalancing (AFFR). We extend the original self-training paradigm to adaptive class-rebalancing self-training (ACRST) based on FBR and AFFR.

We first propose FBR to address the foreground-background imbalance in SSOD. FBR first extracts foreground instances from entire datasets according to ground truths/pseudo labels stored in the CropBank. Thereafter, foreground instances are augmented and pasted to random locations in training images. With synthetic data, FBR can increase the proportion of foreground instances in training targets and alleviate the foreground-background imbalance. 

For foreground-foreground imbalance, we propose adaptive foreground-foreground rebalancing(AFFR) based on FBR. In particular, we design a novel criterion called \emph{pseudo recall} to judge whether a class is neglected or over-focused in SSOD. Thereafter, pseudo-labels of the neglected classes are sampled more frequently because of higher negative confidence. Consequently, the entire dataset is foreground-foreground rebalanced, thereby leading to a minimally biased detector for online pseudo-labeling in the subsequent self-training. 

\begin{figure}[h]
\includegraphics[width=8cm]{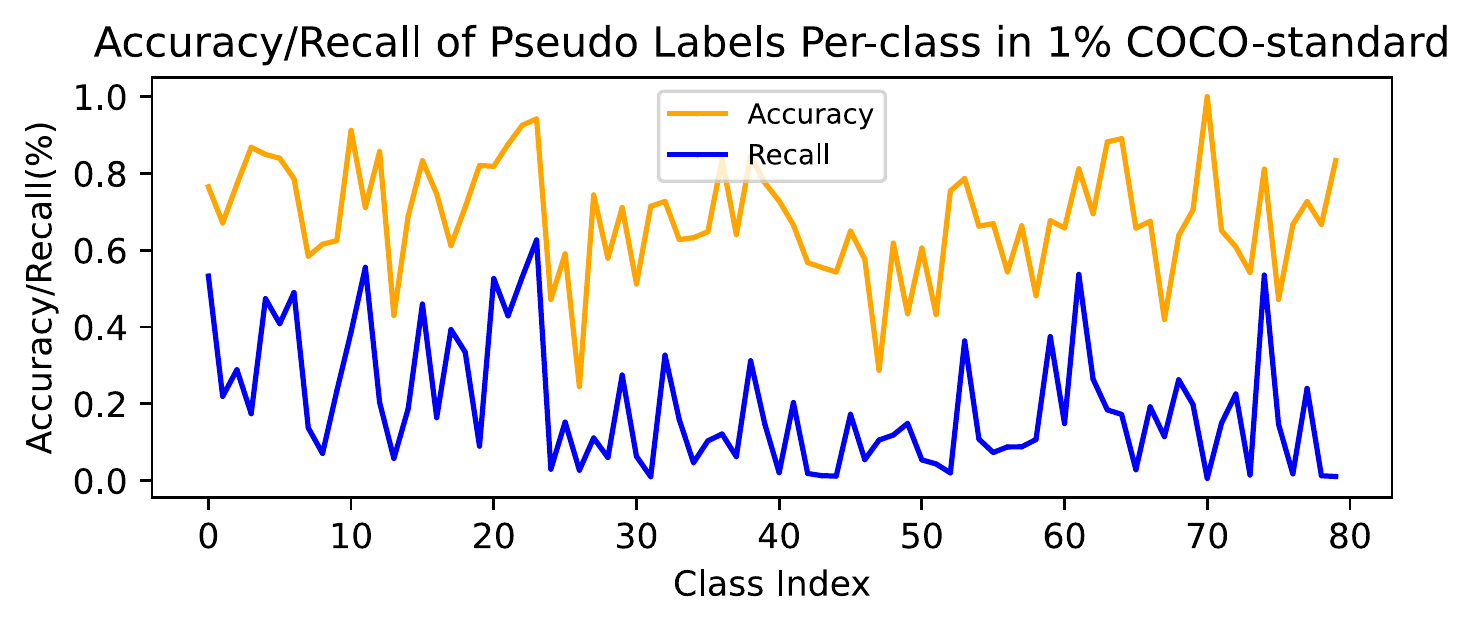}
\caption{\textbf{Accuracy and Recall of pseudo-labels for each class in 1\% COCO-standard.}}
\label{fig:2}
\end{figure}

However, as presented in Figure~\ref{fig:2}, the accuracy of pseudo-labels is undesirable. We observe that FBR and AFFR suffer from noisy pseudo-labels in the mutual-training stage. Therefore, we exploit additional high-level semantics to filter noisy pseudo-labels. In particular, we propose a semi-supervised multi-label classification module to generate image-level pseudo-labels for unlabeled data. Thereafter, we design a two-stage filter mechanism to filter out pseudo-labels that activates negative in classification confidences or image-level pseudo-labels.

Our method outperforms previous state-of-the-art methods on MS-COCO and VOC benchmarks by significant margins. When using only $1\%$ labeled COCO-standard~\cite{2014Microsoft}, our method obtains $5.32$ mAP improvement over state-of-the-arts. When using VOC07~\cite{2010The} as labeled data, our method outperforms state-of-the-arts by $1.43$ mAP improvement.

We summarize our contributions as follows:
\begin{itemize}
    \item We design a novel memory module called CropBank to disentangle fore/background and fore/foreground instances in detection data. With the CropBank, we further propose adaptive class-rebalancing self-training (ACRST) to address the foreground-background and foreground-foreground imbalance in SSOD.
    \item We propose a semi-supervised multi-label classification module to mine high-level semantics from unlabeled data. Thereafter, we propose a two-stage pseudo-label filtering mechanism with classification confidence and high-level semantics. This mechanism is effective in pseudo-label denoising, thereby further facilitating FBR and AFFR.
    \item The proposed data-rebalancing and pseudo-label filtering algorithms are plug-and-play for any self-training-based SSOD framework. Moreover, the CropBank provides an effective detection-specific data augmentation algorithm.
\end{itemize}

\section{Related work}

\subsection{Semi-supervised Learning}
Semi-supervised learning (SSL) utilizes unlabeled data to facilitate model training when large-scale annotated datasets are unavailable. The majority of the SSL methods typically consist of two types: consistency regularization and pseudo-labeling. Consistency regularization~\cite{2019MixMatch,DBLP:journals/corr/abs-1911-09785,xie2020unsupervised,2018Virtual,2016Regularization} enforces the prediction consistency of different augmented views of the same image. Pseudo-labeling~\cite{tarvainen2018mean,2014Learning,DBLP:journals/corr/abs-1908-02983,2019Label} exploits high-quality pseudo-labels of unlabeled data to refine the model pre-trained on few labeled data.SSL has made remarkable success in image classification. However, foreground-background imbalance and foreground-foreground in SSOD heavily impede the effectiveness of the current SSL approaches. To address these limitations, we propose a pseudo-labeling-based  method to alleviate the class imbalance.

\subsection{Semi-supervised Object Detection}
Object detection is a fundamental task in computer vision. Existing object detection frameworks include two- and one-stage detectors. Two-stage detectors~\cite{2017Faster,2017Mask,2013Rich,2015Fast} first generate regions of interest (RoIs) and perform refinement on RoIs thereafter for the final bounding-boxes classification and regression. For one-stage detectors~\cite{2016You,2017Focal,DBLP:journals/corr/abs-1808-01244,0CenterNet}, their predictions are performed on dense grids directly. Although existing studies have made remarkable progress in the past years, they have primarily focused on detectors in a fully supervised setting. SSOD algorithms, which train detectors with a combination of labeled and unlabeled data following standard SSL settings, have increasing attention recently. CSD~\cite{2019csd} utilizes a consistency-based mechanism, which enforces the model predictions consistency of different flipped versions of the same image for generalized feature learning. Similar to CSD, ISD~\cite{2020isd} imposes consistency-regularization on input images and their mixed versions. STAC~\cite{sohn2020detection} introduces a pseudo-labeling-based method, which first pre-trains a detector on available labeled data and generates pseudo-labels on unlabeled data thereafter to re-train the detector. Instant Teaching~\cite{zhou2021instantteaching} develops an SSOD framework with MixUp~\cite{2017mixup} and Mosaic~\cite{DBLP:journals/corr/abs-2004-10934}. Although these studies have improved the performance against the model in a supervised setting, they lack considerations into serious data imbalance issues in SSOD. Recently, Unbiased-Teacher~\cite{liu2021unbiased} has been proposed recently to utilize focal loss~\cite{2017Focal} to alleviate class imbalance. However, the effectiveness of focal loss on unlabeled data is impeded by noisy pseudo-labels. Moreover, Unbiased-Teacher does not facilitate an increase in data diversity. To address the preceding issues and enhance the performance of SSOD, we propose FBR and AFFR to alleviate foreground-background and foreground-foreground imbalance simultaneously. We also devise a two-stage pseudo-label filtering algorithm with classification confidence and high-level semantics.
\begin{figure*}
\includegraphics[width=16cm]{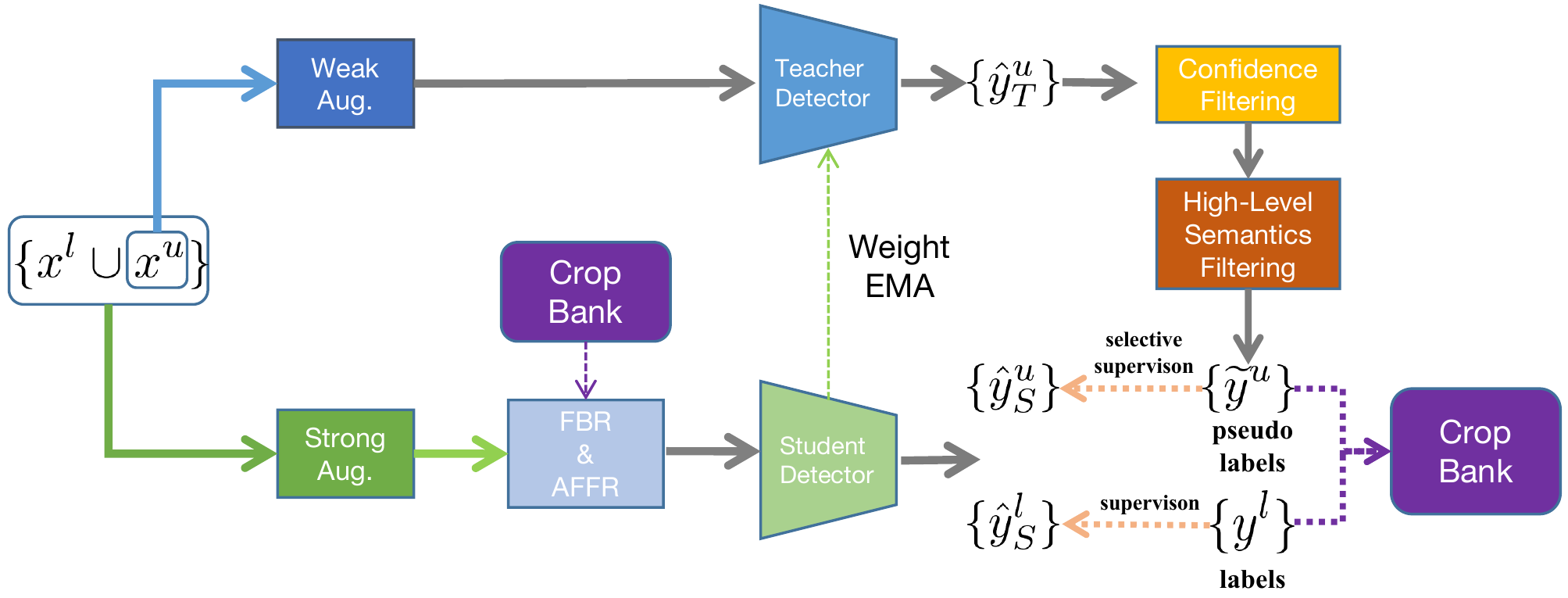}
\caption{\textbf{Overview of our methods}. We take Mean Teacher as our SSOD baseline. In Mean Teacher, the teacher model generates pseudo labels from weakly augmented unlabeled data, and the student model is trained with a combination of ground-truths and pseudo labels. To alleviate the class imbalance in SSOD, we first design a memory module called CropBank which absorbs instance-level annotations including ground truths and pseudo labels. Thereafter, we utilize the CropBank to perform foreground-background rebalancing (FBR) and adaptive foreground-foreground rebalancing (AFFR) on strong augmented unlabeled data for adaptive class-rebalancing self-training (ACRST). To further improve ACRST, we introduce a two-stage pseudo-label filtering algorithm with classification confidence and high-level semantics.}
\label{fig:3}
\end{figure*}

\section{Method}
This section describes our solutions in detail.Section~\ref{3.1} defines the problem, while Section~\ref{3.2} introduces the baseline framework Mean Teacher. Section~\ref{3.3} defines the CropBank, and Section~\ref{3.4} introduces the proposed adaptive class self-training rebalancing (ACRST) algorithm. Section~\ref{3.5} illustrates the two-stage pseudo-label filtering method. Lastly, Section~\ref{3.6} introduces the selective supervision mechanism. The overview of the entire training process is shown in Figure~\ref{fig:3}.

\subsection{Problem Definition}\label{3.1}
Semi-supervised object detection aims to train detectors in a semi-supervised setting, where a small labeled dataset $D_s=\{x_i^s, y_i^s\}_{i=1}^{N_s}$ and a large unlabeled dataset $D_u=\{x_i^u\}_{i=1}^{N_u}$ are available. $N_s$/$N_u$ presents the number of labeled/unlabeled data.$y_i^s$ contains bounding-box annotations including object locations, sizes, and categories in $i$th labeled image $x_i^s$. 

\subsection{Mean Teacher for Semi-supervised Object Detection}\label{3.2}
    This study takes the Mean Teacher~\cite{tarvainen2018mean} as the SSOD baseline. Mean Teacher consists of a teacher and a student model, in which the entire framework is optimized via a mutual learning mechanism. The training pipeline of Mean Teacher consists of two stages. 
    
    \textbf{Pre-training.} 
   The student model is pre-trained with a small amount of labeled data $D_s$ via gradient back-propagation in a supervised manner. Thereafter, we initialize the teacher model with pre-trained model weights of the student model, which produce noisy-less pseudo-labels, thereby facilitating the subsequent training. 
    
    \textbf{Teacher-Student Mutual Learning. } 
    In the mutual learning stage, we train the student model with supervision signals consisting of ground truths and pseudo-labels. Once the student model is updated via the gradient back-propagation, the learned knowledge is feedback to the teacher model in an exponential moving average (EMA) mechanism,
        \begin{equation}
            \theta_s \gets \theta_s + \frac{\partial \mathcal L}{\partial \theta_s},
        \end{equation}
        \begin{equation}
            \theta_t \gets \alpha\theta_t + (1-\alpha)\theta_s,
        \end{equation}
    where $\theta_s/\theta_t$ represents the model parameters of the student/teacher model, and $\mathcal L$ represents the total SSOD losses.
    
    By using Faster-RCNN\cite{2017Faster} as the detection module, the loss function of SSOD can be summarized as a combination of losses on labeled data $\mathcal L_{sup}$ and unlabeled data $\mathcal L_{unsup}$.
        \begin{equation}
        \mathcal L = \mathcal L_{sup} + \lambda_{unsup}\mathcal L_{unsup},
    \end{equation}
    \begin{equation}
        \begin{split}
        \mathcal L_{sup} = \Sigma_i \mathcal L_{cls}^{rpn}(x_i^s,y_i^s) + \mathcal L_{reg}^{rpn}(x_i^s,y_i^s) \\ + \mathcal L_{cls}^{roi}(x_i^s,y_i^s) + \mathcal L_{reg}^{roi}(x_i^s,y_i^s),
        \end{split}
    \end{equation}
    \begin{equation}
        \mathcal L_{unsup} = \Sigma_i \mathcal L_{cls}^{rpn}(x_i^u,\widetilde{y}_i^u) + \mathcal L_{cls}^{roi}(x_i^u,\widetilde{y}_i^u),
    \label{equation_5}
    \end{equation}
where $\mathcal L_{cls}^{rpn}$ is the RPN classification loss,$\mathcal L_{reg}^{rpn}$ is the RPN regression loss, $\mathcal L_{cls}^{roi}$ is the ROI classification loss, $\mathcal L_{reg}^{roi}$ is the ROI regression loss. $y_i^s$ represents the annotation of the labeled image $x_i^s$, $\widetilde{y}_i^u$ represents the pseudo-labels of unlabeled image $x_i^u$, and $\lambda_{unsup}$ is used to balance the supervised and unsupervised losses. Note that regression losses are removed in $L_{unsup}$ in previous SSOD studies for denoising.

To succeed in SSOD, the teacher model must generate accurate pseudo-labels and maintain a reliable performance margin over the student model throughout the training. However, we observe that class imbalance in SSOD significantly hinders the performance of the teacher model.

\subsection{CropBank}\label{3.3}

Data-rebalancing algorithms have been proved to be the most simple and effective data-rebalancing method in classification tasks. However, their effectiveness is heavily impeded by strong interconnections on both foreground-background and foreground-foreground instances. To separate the entanglement, we propose a novel memory module called CropBank, which stores abundant instance-level annotations. The CropBank consists of two sub-banks, namely, Labeled CropBank $\Phi_L=\{y_i^l\}_{i=1}^{N_L}$ and Pseudo CropBank $\Phi_U=\{\widetilde{y}_i^u\}_{i=1}^{N_U}$, where $N_L$/$N_U$ represent the size of Labeled/Pseudo CropBank, $y_i^l$/ $\widetilde{y}_i^u$ represents ground truths/pseudo-labels of $i$th labeled/unlabeled image.

In the implementation, the CropBank size is unlimited owing to the negligible memory consumption of instance-level annotations.  In the training stage, $\Phi_L$ is fixed once generated, while $\Phi_U$ is updated periodically with improved pseudo-labels in mutual training. We use CropBank as the basis to decouple instances and design adaptive class-rebalancing self-training (ACRST) to address the class imbalance. 

\begin{figure}[h!]
\includegraphics[width=8.5cm]{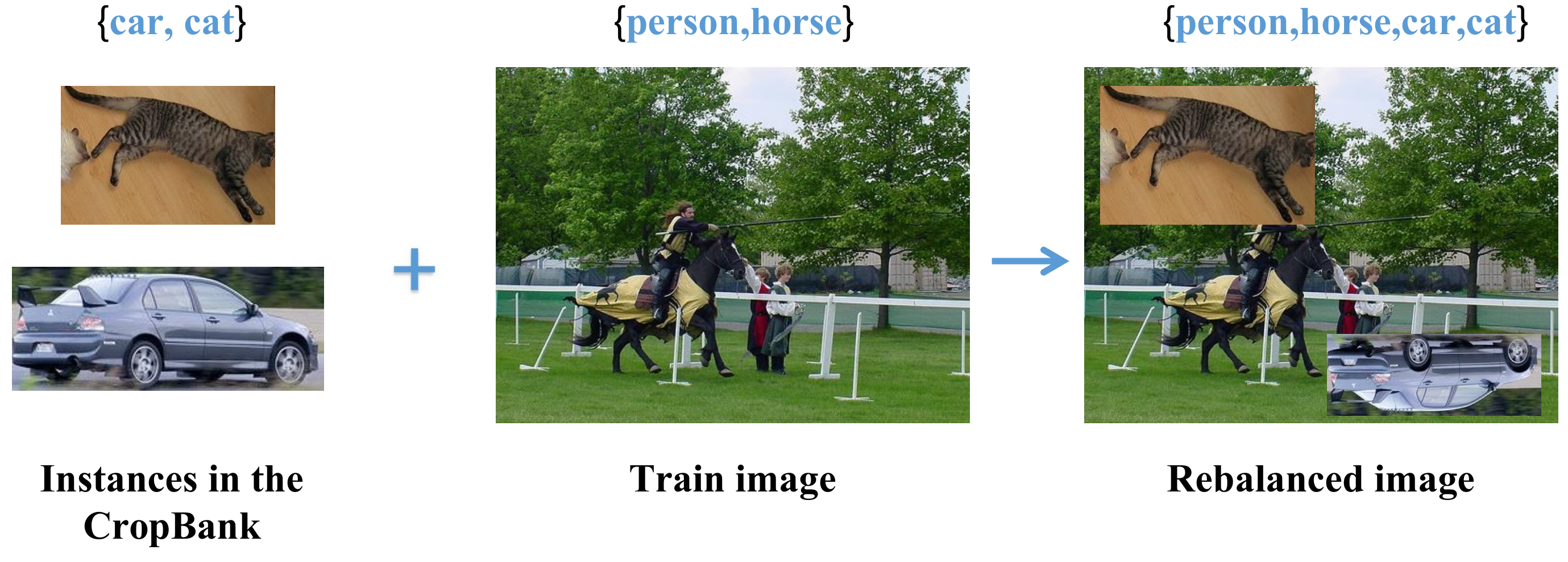}
\caption{\textbf{Data-rebalancing with instances in the CropBank.}}
\label{fig:4}
\end{figure}

\subsection{Adaptive Class-Rebalancing Self-Training}\label{3.4}

The self-training paradigm is an ideal solution to alleviate the lack of labeled data. However, its effectiveness is impeded by the inherent class imbalance in object detection tasks. Therefore, we propose ACRST to alleviate the class imbalance in SSOD. ACRST consists of foreground-background rebalancing (FBR) and adaptive foreground–foreground rebalancing (AFFR).

\subsubsection{Foreground-Background Rebalancing} \label{3.4.1}
\ 
\newline
Models trained on foreground-background imbalanced data tend to overfit excessive background instances. Foreground-background imbalance in object detection has been widely explored.  Various solutions have been proposed to alleviate such an imbalance, including loss-reweighting~\cite{2017Focal} and region refinement~\cite{2017Faster}. Unfortunately, these methods rely on ground truths unavailable in SSOD to guide the rebalancing procedure. Hence, we utilize abundant instance-level annotations including ground truths and pseudo-labels in the CropBank to perform foreground-background rebalancing. 

Given a training mini-batch $B=\{x_i, y_i\}_{i=1}^{N_T}$, we fetch a set of foreground instances $F=\{c_j, y_j\}_{j=1}^{N_C}$ from the CropBank $\Phi_L$ and $\Phi_U$ for each image $x_i$ following a sampling distribution \emph{P}, where $c_j$ is a foreground instance cropped from original images with annotation $y_j$. Thereafter, a new training mini-batch  $B_{mix}=\{x_i^{mix}, y_i^{mix}\}_{i=1}^{N_{T}}$ is generated as follows:
    \begin{equation}
        x_i^{mix} = \alpha x_i + (1-\alpha) c_j,
    \end{equation}
    \begin{equation}
        y_i^{mix} = merge(y_i, y_j).
    \end{equation}    
Where $\alpha$ denotes a binary mask of pasted objects and $y_i^{mix}$ denotes mixed annotations, in which fully occluded instances are removed from mixed image $x_i^{mix}$. In detail, $c_j$ is a rectangular region cropped from the image based on instance-level annotations in the CropBank. During training, $c_j$ is augmented and pasted thereafter to random locations of $x_i$. This combined procedure increases the ratio of foreground instances in the training data for foreground-background rebalancing and also explores essential context semantics from a holistic perspective. 

Once mixed images are ready, we take them to train the detector as the pipeline of Mean Teacher. The rebalancing process is shown in Figure~\ref{fig:4}. As discussed in Section~\ref{3.6}, such a crop-and-paste operation enables higher model performance with selective supervision. 

\subsubsection{Adaptive Foreground-Foreground Rebalancing}\label{3.4.2}
\ 
\newline
FBR adequately alleviates the foreground-background imbalance with considerable attention on foreground instances. However, a random sampling distribution \emph{P}, such as uniform distribution, fails to correct the foreground-foreground imbalance. Hence, we propose an adaptive sampling probability distribution \emph{P} for foreground-foreground rebalancing. In particular, samples in neglected classes are selected more frequently during the training.

To measure the neglected degree of a class, we propose a novel criterion \emph{pseudo recall} ($PR$), which quantities the proportion of pseudo-labels to ground truths. In detail, we estimate the class distribution of unlabeled data from labeled data on account of distribution similarity between labeled and unlabeled data. Suppose there are $K$ classes $\{1,2,..K\}$ in datasets. We calculate \emph{pseudo recall} for class $k$ as
    \begin{equation}
        PR_k = \frac{N_k^{u}}{r N_k^{l}},
    \end{equation}
where $N_k^{u}$ and $N_k^{l}$ denote the number of pseudo-labels and ground truths of class $k$, and $r$ is the ratio of the unlabeled to labeled data. 

\emph{Pseudo recall} defines how neglected one class is under the SSOD setting. High $PR_k$ indicates that the detector is certain even overconfident on class $k$. Consequently, lower sampling probabilities should be allocated to samples in class $k$ for overfitting alleviation. By contrast, low $PR_k$ implies that the detector lacks confidence for detecting instances of class $k$. Therefore, we should select these instances frequently. As a solution, we sort the classes in descending order according to \emph{pseudo recall} and design the following adaptively sampling strategy:
    \begin{equation}
        \mu_k = \left (\frac{PR_{K-k+1}}{\Sigma_{i=1}^K PR_i} \right )^\beta,
    \end{equation}
where $\mu_k$ is the probability of selecting instances of class $k$, and $\beta$ is used to tune the sampling probability. This mechanism adaptively allocates higher/lower sampling rates to neglected/over-focused instances. Note that AFFR performs FBR simultaneously. There are numerous ways to rebalance class distribution, and we introduce an effective example. A potential problem with this mechanism is that the noise of pseudo-labels in the neglected classes is amplified. Therefore, we propose a two-stage pseudo-label filtering mechanism in Section~\ref{3.5}.

\subsection{Two-stage Pseudo-label Filtering}\label{3.5}

The proposed ACRST considerably alleviates the class imbalance in SSOD. However, its effectiveness is heavily affected by the quality of pseudo-labels. Once noise in the CropBank is selected improperly, it will be undesirably amplified in self-training. Consequently, we should filter noisy pseudo-labels from the teacher model predictions and store noisy-less pseudo-labels in the CropBank. In SSOD, the general filtering algorithm sets a threshold $\tau_{cls}$ to filter predictions with low classification confidence out. However, such single-stage filtering without additional semantics constraints is prone to produce noisy pseudo-labels. As a solution, we propose a semi-supervised multi-label classification module to learn high-level semantics (i.e., image-level pseudo-labels). Thereafter, we design a two-stage filtering algorithm with classification confidences and high-level semantics to generate accurate pseudo-labels.

\subsubsection{Semi-supervised Multi-label Classification.} 
\ 
\newline
The proposed semi-supervised multi-label classification module is devised based on Mean Teacher for the classification task. For each image $x_i$, we particularly aim to predict its image-level pseudo-labels $v_i=\{l_k\}_{k=1}^{K}, l_k \in \{0,1\}$, where $K$ is the total category number and $l_k$ determines whether there are instances of class $k$ in the image. In the training stage, predictions of the teacher model are converted to image-level pseudo-labels which supervise the student model. We utilize a focal–binary–cross-entropy loss to optimize the student model.

\subsubsection{Two-stage Pseudo-label Filtering. } 
\ 
\newline
For bounding-box prediction $i$ with classification score $s_i$ of image $x_j$ with image-level pseudo-label $y_j$, we perform a two-stage filtering to get bounding-box pseudo-labels with low-noise. In the first stage, we filter predictions out with scores $s < \tau_{cls}$ to remove predictions with low objectness or wrong class labels. In the second stage, predictions whose classes activate negative in $y_j$ (i.e., activation values are smaller than $\tau_{ml}$) are removed. The second filtering stage utilizes high-level semantics to filter noisy predictions inconsistent with image-level pseudo-labels. With the two-stage filtering, the consistency of low- and high-level semantics are achieved. 

\subsection{Selective Supervision}\label{3.6}
In previous SSOD research, bounding-box regression losses are removed during training to alleviate noise. By contrast, utilizing regression losses in our framework is beneficial to achieve high SSOD performance, which is attributed to the CropBank module.
    
The contribution is two-fold. First, the CropBank alleviates noise from partially detected instances, which take a large proportion in bias predictions. Learning blindly with these noisy pseudo-labels will heavily aggravate the model performance.  However, when the partially detected bounding-boxes from the CropBank are cropped and pasted to training batches, they become independent and complete in the new background, thereby providing additional clean training supervisions. Second, the CropBank provides a detection-specific data augmentation method. The additional augmented data continuously improves the regression accuracy of pseudo-labels.

With selective supervision, loss function $\mathcal L_{unsup}$ in Equation \ref{equation_5} can be represented as follows:
\begin{equation}
    \begin{split}
    \mathcal L_{unsup} = \Sigma_i \mathcal L_{cls}^{rpn}(x_i^u,\widetilde{y}_i^u) + \mathcal L_{reg}^{rpn}(x_i^u,\widetilde{y}_i^{ss}) \\ + \mathcal L_{cls}^{roi}(x_i^u,\widetilde{y}_i^u) + \mathcal L_{reg}^{roi}(x_i^u,\widetilde{y}_i^{ss}),
    \end{split}
\end{equation}
where $\widetilde{y}_i^{ss}$ denotes instances from the CropBank in $x_i^u$.

\section{Experiments}

\begin{table*}[h!]  
\centering
\caption{Experimental results on COCO-standard comparing with CSD, STAC and Unbiased Teacher.} 
\begin{tabular}{cccccc}
     \hline
      &  \multicolumn{5}{c}{COCO-standard ($AP_{50:95}$)} \cr
     \cmidrule(lr){2-6}
     & 0.5\% & 1\% & 2\% & 5\% & 10\% \\
     \hline
     Supervised & $6.83\pm0.15$ & $9.05\pm0.16$ & $12.70\pm0.15$ & $18.47\pm0.22$ & $23.86\pm0.81$ \\
     CSD\cite{2019csd} & $7.41\pm0.21$ & $10.51\pm0.06$ & $13.93\pm0.12$ & $18.63\pm0.07$ & $22.46\pm0.08$ \\
     STAC\cite{sohn2020detection} & $9.78\pm0.53$ & $13.97\pm0.35$ & $18.25\pm0.25$ & $24.38\pm0.12$ & $28.64\pm0.21$ \\
     Instant Teaching\cite{zhou2021instantteaching} & - & $18.05\pm0.15$ & $22.45\pm0.15$ & $26.75\pm0.05$ & $30.40\pm0.05$ \\
     Unbiased Teacher\cite{liu2021unbiased} & $16.94\pm0.23$ & $20.75\pm0.12$ & $24.30\pm0.07$ & $28.27\pm0.11$ & $31.5\pm0.10$ \\
     Ours & $\textbf{19.62} \pm \textbf{0.37}$ & $\textbf{26.07}\pm\textbf{0.46}$ & $\textbf{28.69} \pm \textbf{0.17}$ & $\textbf{31.35} \pm \textbf{0.13}$ & $\textbf{34.92} \pm \textbf{0.22}$  \\
     \hline
\label{table:1}
\end{tabular}
\end{table*}

\begin{table*}[h!]
    \centering
    \caption{Experimental results on COCO-additional with CSD, STAC, and Unbiased Teacher. Note that N×represents N×90K training iterations.}
    \begin{tabular}{cccccc}
     \hline
        &  \multicolumn{5}{c}{COCO-additional ($AP_{50:95}$)}   \cr
     \cmidrule(lr){2-6}
        &Supervised(3×)&CSD(3×) 
        &STAC(6×) 
        &Unbiased Teacher(3×)
        & Ours(3×)\\
     \hline
      $AP_{50:95}$ & $40.20$ & $38.82$ & $39.21$ & $41.30$ & $\textbf{42.79}$  \\
    \hline
    \end{tabular}
    \label{table:2}
\end{table*}

\subsection{Datasets} \label{dataset}
We evaluate our method on three SSOD benchmarks from MS-COCO~\cite{2014Microsoft} and PASCAL VOC~\cite{2010The}. 
\begin{enumerate}
    \item \emph{COCO-standard}: We sample $0.5$/$1$/$2$/$5$/$10\%$ of the COCO2017-train set as the labeled dataset and take the remaining data as the unlabeled dataset.
    \item \emph{COCO-additional}: We use the COCO2017-train set as the labeled dataset and the additional COCO2017-unlabeled set as the unlabeled dataset. 
    \item \emph{VOC07\&12}: We use the VOC07-trainval set as the labeled dataset and the VOC12-trainval set as the unlabeled dataset. 
\end{enumerate}
We evaluate the model performance on the COCO2017-val set for (1)(2) and VOC07-test set for (3).

\subsection{Implementation Details} \label{impementation}
We use FPN-Faster-RCNN with ResNet-50 backbone as the detection module. ResNet-50 is initialized with ImageNet pre-trained weights. We set the hyper-parameters $\lambda_{unsup}=2$, $\beta=2$.  For two-stage pseudo-label filtering, we use classification confidence threshold $\tau_{cls}=0.7$ and multi-label confidence threshold $\tau_{ml}=0.2$. We use $AP_{50:95}$, i.e, mAP as the evaluation metric. We construct each training batch with $32$ labeled and $32$ unlabeled images for all the training settings. For the \emph{COCO-standard}, the pre-training stage takes $2500$/$5000$/$10000$/$20000$/$40000$ steps for $0.5$/$1$/$2$/$5$/$10\%$ \emph{COCO-standard} and $180000$ steps for the whole training stage of \emph{COCO-standard}. For \emph{COCO-additional}, the pre-training stage takes $90000$ steps in total $270000$ training steps. For \emph{VOC07\&12}, the pre-training stage takes $12000$ steps and $36000$ steps for the entire training stage. Strong augmentations in our research consist of random jittering, Gaussian noise, and random crop. Weak augmentations in our study consist of random resize and flip. Moreover, we set the above hyper-parameters without aggressively searching. Consequently, high model performance may be achieved with improved choices.

\subsection{Results and Comparisons} \label{results}
\subsubsection{COCO-standard}
\ 
\newline
We ﬁrst evaluate the efﬁcacy of our method on COCO-standard. As shown in Table~\ref{table:1}, when only $0.5\%$ to $10\%$ of the entire dataset are labeled, our model consistently performs better against all previous studies in CSD, STAC, Instant Teaching, and Unbiased Teacher. When trained on the $1\%$ COCO-standard, our method achieves $5.32$ mAP improvement compared Unbiased-Teacher. The mAP is even higher than CSD trained on $10\%$ COCO-standard. When trained on $10\%$ COCO-standard, our method achieves $10.42$ mAP improvement compared with supervised baselines. We attribute the success of model performance to two factors.

\textbf{Class rebalanced data}. Our method alleviates the class imbalance in SSOD with two rebalancing algorithms (i.e., FBR and AFFR). Foreground-background rebalanced data prevents the model from overfitting on background instances and helps mine beneficial information from enormous unlabeled data. Foreground-foreground rebalanced data benefits the model predictions with information from neglected classes and avoids biased predictions on over-focused classes. 

\textbf{Noise-less pseudo-labels}. When using the self-training paradigm for SSOD, accurate and reliable pseudo-labels from pre-trained models should be generated. We propose a teacher–student mutual learning mechanism for progressive pseudo-labels refinement. In addition, we introduce a two-stage pseudo-label filtering algorithm to remove noisy predictions with classification confidence and high-level semantics. With accurate pseudo-labels, the student model is well optimized and gives beneficial feedback to the teacher model. We present an ablation study on two-stage pseudo-label filtering in Section~\ref{ablationstudy}.


\subsubsection{COCO-additional}
\ 
\newline
In this section, we verify whether our method can further improve the model trained on a large-scale labeled dataset with additional unlabeled data. Table \ref{table:2} shows that our model has a $0.41$ mAP improvement compared with those of previous methods, and $1.51$ mAP improvement compared with the supervised baseline. This result indicates that our method achieves satisfying improvement even on the well-trained model.

\begin{table*}[h!]  
\centering
\caption{Experimental results on VOC07\&12 comparing with CSD, STAC and Unbiased Teacher.} 
\begin{tabular}{cccccc}
     \hline
      & Labeled & Unlabeled & $AP_{50}$ & $AP_{50:95}$ \\
    \hline
    Supervised & VOC07 & None & $72.63$ & $42.13$ \\
    \hline
    CSD\cite{2019csd} & VOC07 & VOC12 & $74.70$ & - \\
    STAC\cite{sohn2020detection} & VOC07 & VOC12 & $77.45$ & $44.64$ \\
    Unbiased Teacher\cite{liu2021unbiased} & VOC07 & VOC12 & $77.37$ & $48.69$ \\
    Ours & VOC07 & VOC12 & $\textbf{78.16}$ & $\textbf{50.12}$ \\
    \hline
\end{tabular}
\label{table:3}
\end{table*}

\subsubsection{VOC07\&12}
\ 
\newline
We evaluate models on a less imbalanced dataset VOC07\&12 to demonstrate the generalization of our method. Table \ref{table:3} provides the mAP results of CSD, STAC, Unbiased Teacher, and our method. Our method achieves $7.99$ mAP improvement compared with the supervised setting, and $1.43$ mAP improvement against previous state-of-the-art methods, even though Unbiased Teacher has witnessed performance saturation in VOC07\&12. We owe the success of our approach to the generalization ability of ACRST. Even if training data is foreground-foreground balanced, FBR can substantially alleviate the inevitable foreground-background imbalance in SSOD. Furthermore, the two-stage pseudo-label filtering mechanism benefits the model trained on VOC07\&12.

\begin{table}[h!]  
\centering
\caption{Ablation study in 1\% \emph{COCO-standard}.} 
\begin{tabular}{ccccc}
     \hline
     FBR & AFFR & Two-Stage & SS & $AP_{50:95}$ \\
    \hline
     &  &   & & 20.75 \\
     &  &  \checkmark &  & $23.48$ \\
    \checkmark &  &   & & $23.32$ \\
    \checkmark & \checkmark & & &$24.17$ \\
    \checkmark & \checkmark& \checkmark &  & $25.56$ \\
    \checkmark & \checkmark& \checkmark & \checkmark & $\textbf{26.12}$ \\
    \hline
\label{table:4}
\end{tabular}
\end{table}

\begin{figure*}[h!]
    \centering
    \begin{minipage}{8cm}
      \centerline{\includegraphics[width=8.0cm]{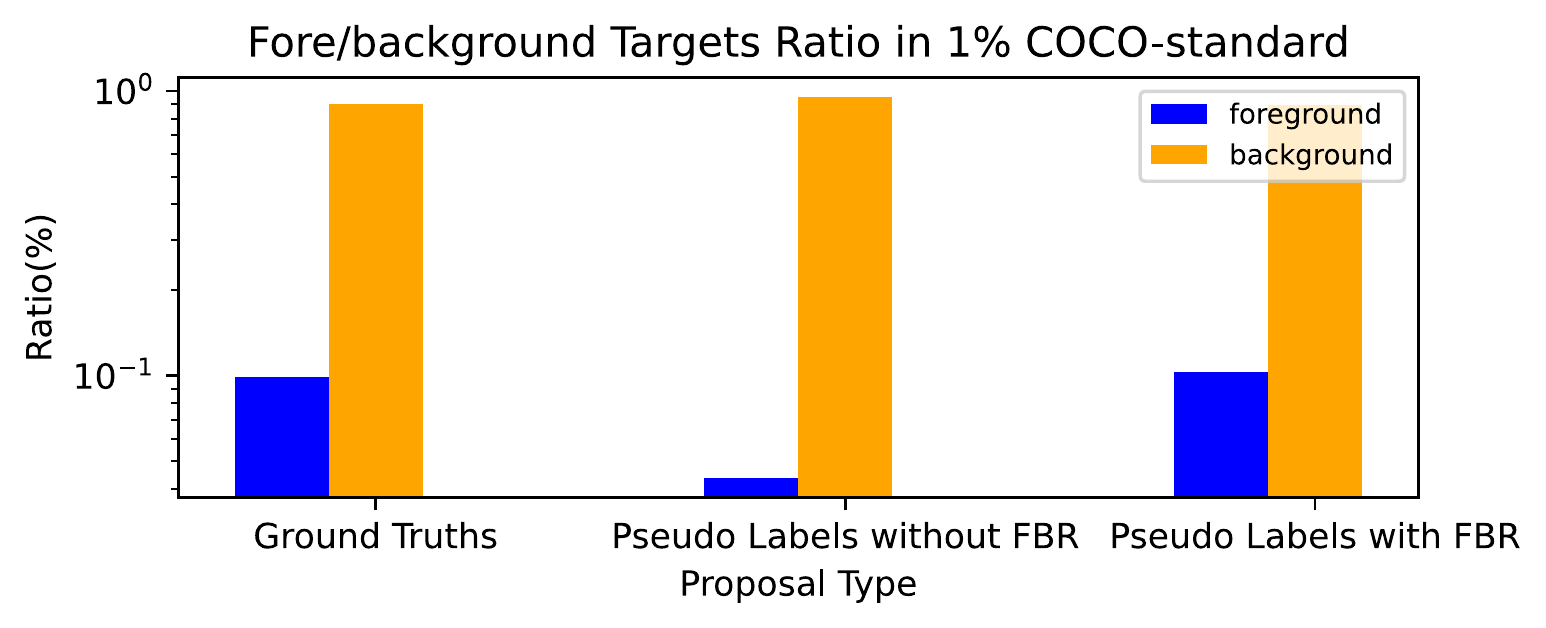}}
        \centerline{(a)}
    \end{minipage}
    \begin{minipage}{8cm}
      \centerline{\includegraphics[width=8.0cm]{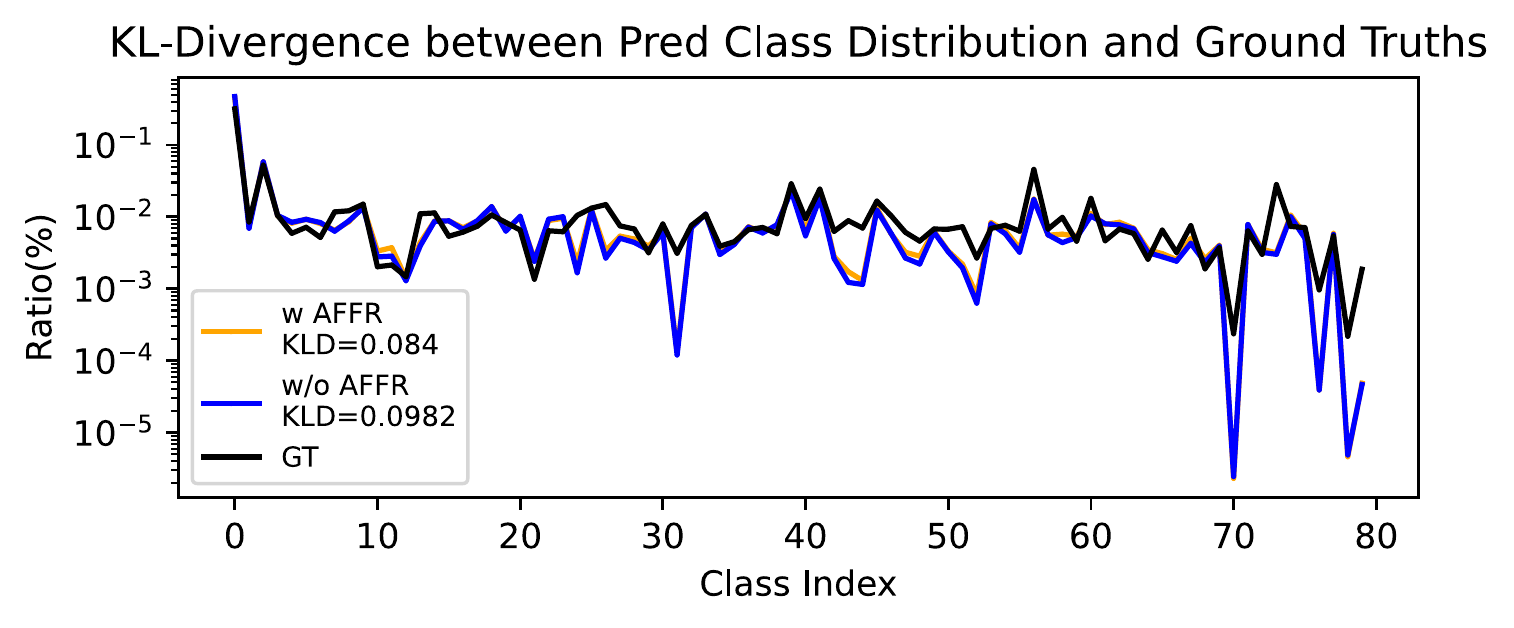}}
        \centerline{(b)}
    \end{minipage}
    \caption{Ablation study on the FBR (a) and AFFR (b). (a) FBR alleviates foreground-background imbalance in pseudo-labels in 1\% COCO-standard. (b) AFFR reduces the KL-Divergence(KLD) between the pseudo-labels distribution and the ground truths distribution from 0.0982 to 0.084 in 1\% COCO-standard.}
    \label{fig:5}
\end{figure*}

\begin{figure*}[h!]
    \centering
    \begin{minipage}{8cm}
      \centerline{\includegraphics[width=8.0cm]{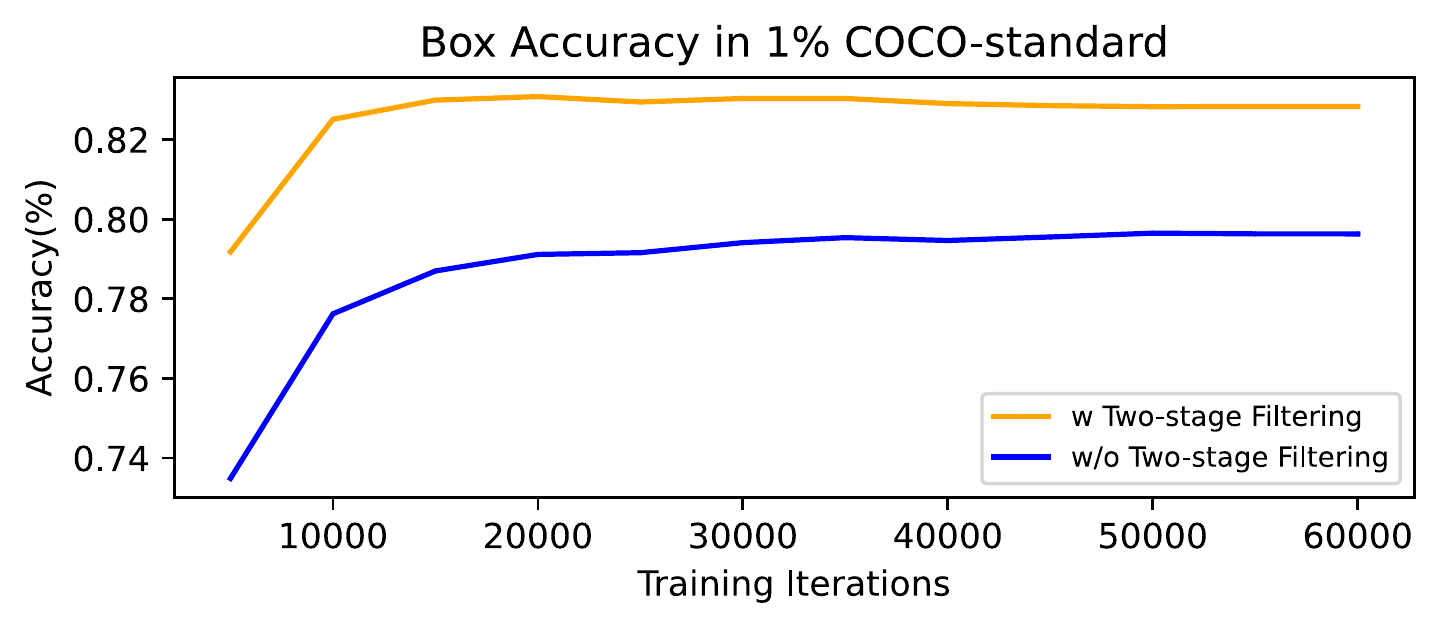}}
        \centerline{(a)}
    \end{minipage}
    \begin{minipage}{8cm}
      \centerline{\includegraphics[width=8.0cm]{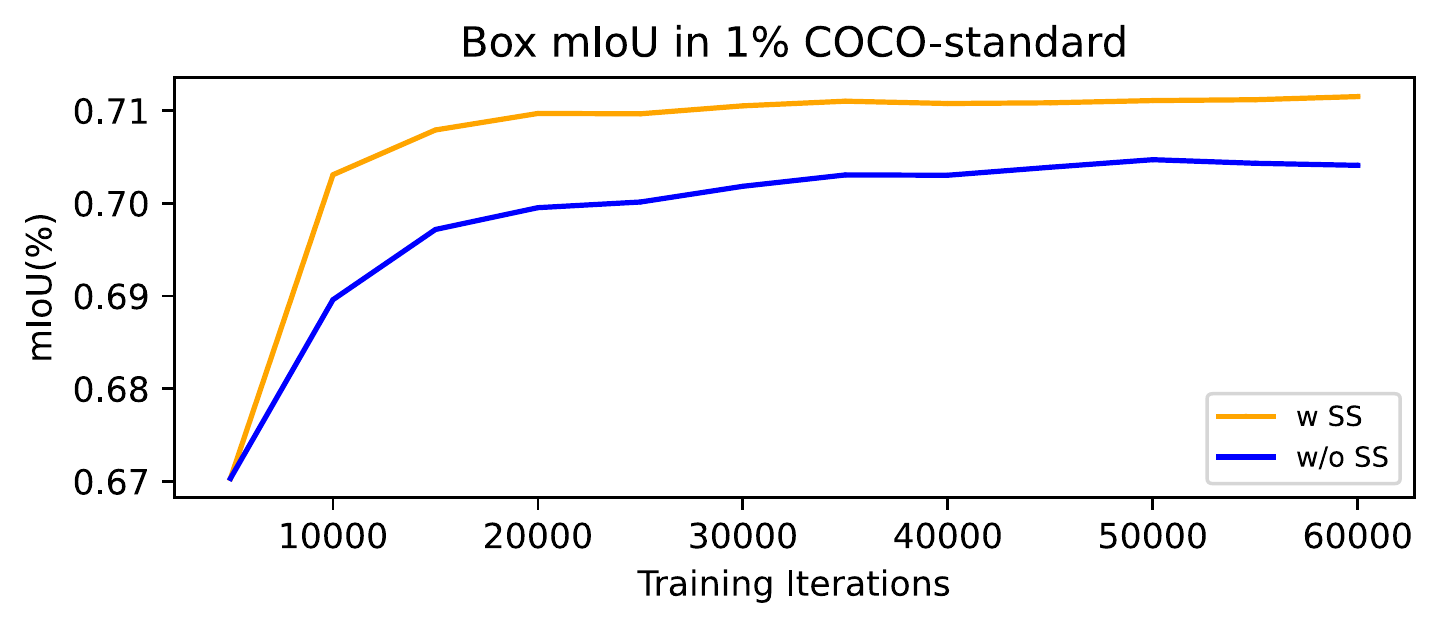}}
        \centerline{(b)}
    \end{minipage}
    \caption{Pseudo-labels improvement on Box Accuracy and Box mIoU in 1\% COCO-standard. (a) Box accuracy of pseudo-labels with/without two-stage pseudo-label filtering. (b) Box mIoU between pseudo-labels and ground truths with/without selective supervision (SS).}
    \label{fig:6}
\end{figure*}

\subsection{Ablation Study} \label{ablationstudy}

\subsubsection{Foreground–Background Rebalancing (FBR)}
\ 
\newline
We first verify the effect of FBR. Table~\ref{table:4} shows that applying FBR improves mAP in $1\%$ labeled COCO from $20.75$ to $23.32$. To analyze the divergent results, we visualize the foreground-background distribution of the rebalanced pseudo-labels. Figure~\ref{fig:5}(a) shows that after rebalancing, the distribution of the foreground and background instances is rebalanced. The ratio of foreground instances in rebalanced pseudo-labels is even higher than that of ground truths. Therefore, training detectors with rebalanced training data alleviates data bias and produces high mAP.

\subsubsection{Adaptive Foreground–Foreground Rebalancing (AFFR)}
\ 
\newline
Table~\ref{table:4} shows that AFFR improves mAP from $23.32$ to $24.17$ based on FBR. We verify the effectiveness of AFFR by analyzing the foreground class distribution. Figure~\ref{fig:5}(b) presents that AFFR alleviates foreground–foreground imbalance and reduces KL-divergence from 0.0982 to 0.084.  This result confirms the effectiveness of AFFR in handling foreground–foreground imbalance issues in pseudo-labels. Using AFFR can generate an unbiased training data distribution and results in a higher mAP.

\subsubsection{Two-stage Pseudo-label Filtering}
\ 
\newline
We also verify the effectiveness of the two-stage pseudo-label filtering with classification confidences and high-level semantics. As presented in Table~\ref{table:4}, the model that filters pseudo-labels with high-level semantics can favorably improve model performance against the model with only classification confidence. The two-stage filtering mechanism utilizes an uncertainty mechanism where only predictions with high objectness and follow image-level constraints are regarded as accurate pseudo-labels. Figure \ref{fig:6}(a) shows that the two-stage filtering mechanism has a continuous improvement on the accuracy of pseudo-labels. This confirms the effectiveness of the two-stage filtering mechanism in removing noisy predictions in SSOD. Moreover, a two-stage filtering mechanism is necessary to build a clean Pseudo CropBank and further improve the performances of ACRST. Table~\ref{table:4} indicates that applying the two-stage filtering mechanism improves the mAP from $20.75$ to $23.48$. Moreover, applying the mechanism further improves the mAP of the model trained with ACRST from $24.17$ to $25.56$. This result confirms that the two-stage filtering mechanism is effective in handling the noisy pseudo labels.

\subsubsection{Selective Supervision (SS)}
\ 
\newline
Lastly, we examine the effectiveness of selective supervision in SSOD. As presented in Table~\ref{table:4}, the selective supervision improves the mAP from $25.56$ to $26.12$ in $1\%$ COCO-standard. We owe the improvement to the crop-and-paste operation in ACRST, in which incomplete instances are pasted to a new background in the training data. As a result, utilizing these instances as targets of regression optimization produces less noise compared to regress them directly in originating images. We further analyze the distribution of regression accuracy of pseudo-labels. Figure~\ref{fig:6}(b) shows that selective supervision improves the mIoU of pseudo-labels. Accordingly, transferring these incomplete predictions to complete objects in a new background alleviates noise in the regression targets and improves the model performance. Selective supervision is still under exploration. For example, the current strategy fails to handle noise in which objects are overlapped with each other.

\section{Discussions}

\subsection{CropBank} Data augmentations that generate different views of the same image are necessary to pseudo-labeling-based and consistency-regularization-based semi-supervised learning. Conventional data augmentations consist of color jittering~\cite{DBLP:journals/corr/abs-1911-09785}, rotation~\cite{2019csd}, and Gaussian Noise~\cite{2020A}. Although these methods are effective in vision tasks, they fail to change the image semantics. MixUp~\cite{2017mixup} and Mosaic~\cite{DBLP:journals/corr/abs-2004-10934} are proposed to change the image semantics by mixing images.However, they fail to decouple instances semantics in detection data.

The CropBank provides a detection-specific data augmentation that effectively decouples entangled semantics in images. The idea of the CropBank is inspired by the CutMix~\cite{0CutMix}. 

The difference between the CropBank and the CutMix is two-fold. First, the CropBank decouples instances in detection data and creates training new detection datasets with complex disentangled semantics, while the CutMix is classification-specific and unable to decouple semantics. Second, the CropBank adaptively injects semantics from the entire dataset to training images, but the CutMix only exchanges image-to-image semantics. Table~\ref{table:5} provides the model performance with different data augmentations in SSOD. The CropBank improves $AP_{50:95}$ from $16.00$ to $16.85$ compared to MixUp and Mosaic in \cite{zhou2021instantteaching}. 

Moreover, the CropBank can be easily embedded in semantic and instance segmentation with few modifications. We will release related researches in future studies.

\begin{table}[h!]  
\centering
\caption{STAC\cite{sohn2020detection} performance in 1\% COCO-standard under different data augmentations.} 
\begin{tabular}{ccc}
     \hline
    Augmentations & $AP_{50:95}$ \\
    \hline
    MixUp and Mosaic~\cite{zhou2021instantteaching} & $16.00$ \\
    CropBank  & $\textbf{16.85}$ \\
    \hline
\end{tabular}
\label{table:5}
\end{table}

\begin{table}[h!]  
\centering
\caption{Model Performance, Accuracy and Recall of pseudo-labels in 1\% COCO-standard in three settings.} 
\begin{tabular}{cccccc}
     \hline
    Setting & Accuracy & Recall  & $AP_{50:95}$ \\
    \hline
    One-stage & $0.735$ & $0.299$ & $20.75$ \\
    Two-stage Filtering & $\textbf{0.792}$ & $0.292$ & $\textbf{23.48}$ \\
    Two-stage Mining & $0.722$ & $\textbf{0.376}$ & $21.56$ \\
    \hline
\end{tabular}
\label{table:6}
\end{table}

\subsection{Image-level Pseudo-labels} There is an unexplored question in our study: Why we utilize high-level semantics (i.e., image-level pseudo-labels) to filter noisy pseudo-labels instead of mining neglected predictions? To answer this question, we first evaluate the model trained in three different settings. (1) \textbf{One-stage:} Predictions with low classification confidence are filtered. (2) \textbf{Two-stage filtering:} Predictions with low classification confidence or low activation in image-level pseudo-labels are filtered. (3) \textbf{Two-stage Mining:} Predictions with high classification confidence or high activation in image-level pseudo-labels are reserved. Thereafter, we calculate the $AP_{50:95}$, accuracy, and recall of pseudo-labels. 

Table~\ref{table:6} indicates that the two-stage mining mechanism improves recall from $0.299$ to $0.376$ and reduces accuracy from $0.735$ to $0.722$, compared to the one-stage strategy.  The two-stage filtering mechanism improves the accuracy from $0.735$ to $0.792$ and reduces the recall from $0.299$ to $0.292$ compared to the one-stage filtering. Although the recall improvement of the two-stage mining is higher than the accuracy improvement of the two-stage filtering, the $AP_{50:95}$ improvement of the latter is $2.73$, which is higher compared to the former. 

This result indicates that a higher accuracy of pseudo-labels is considerably more important than a higher recall. Hence, future studies should carefully consider the balance between accuracy and recall. We will further analyze this problem in future studies.

\section{Conclusion}
This study proposes ACRST based on a novel memory module called CropBank to address the class imbalance in SSOD. ACRST considerably alleviates foreground-background and foreground-foreground imbalance with proposed FBR and AFFR. To further improve FBR and AFFR, we design a two-stage pseudo-label filtering algorithm with classification confidence and high-level semantics. Over iterations on rebalanced training data, SSOD detectors become unbiased and ameliorate the model performance progressively. Extensive experiments on benchmarks demonstrate the effectiveness of our method.

\bibliography{Formatting-Instructions-LaTeX-2021} 

\begin{thebibliography}{38}
\providecommand{\natexlab}[1]{#1}
\providecommand{\url}[1]{\texttt{#1}}
\providecommand{\urlprefix}{URL }
\expandafter\ifx\csname urlstyle\endcsname\relax
  \providecommand{\doi}[1]{doi:\discretionary{}{}{}#1}\else
  \providecommand{\doi}{doi:\discretionary{}{}{}\begingroup
  \urlstyle{rm}\Url}\fi

\bibitem[{Arazo et~al.(2019)Arazo, Ortego, Albert, O'Connor, and
  McGuinness}]{DBLP:journals/corr/abs-1908-02983}
Arazo, E.; Ortego, D.; Albert, P.; O'Connor, N.~E.; and McGuinness, K. 2019.
\newblock Pseudo-Labeling and Confirmation Bias in Deep Semi-Supervised
  Learning.
\newblock \emph{CoRR} abs/1908.02983.
\newblock \urlprefix\url{http://arxiv.org/abs/1908.02983}.

\bibitem[{Bachman, Alsharif, and Precup(2014)}]{2014Learning}
Bachman, P.; Alsharif, O.; and Precup, D. 2014.
\newblock Learning with Pseudo-Ensembles.
\newblock In Ghahramani, Z.; Welling, M.; Cortes, C.; Lawrence, N.~D.; and
  Weinberger, K.~Q., eds., \emph{Advances in Neural Information Processing
  Systems 27: Annual Conference on Neural Information Processing Systems 2014,
  December 8-13 2014, Montreal, Quebec, Canada}, 3365--3373.
\newblock
  \urlprefix\url{https://proceedings.neurips.cc/paper/2014/hash/66be31e4c40d676991f2405aaecc6934-Abstract.html}.

\bibitem[{Berthelot et~al.(2019{\natexlab{a}})Berthelot, Carlini, Cubuk,
  Kurakin, Sohn, Zhang, and Raffel}]{DBLP:journals/corr/abs-1911-09785}
Berthelot, D.; Carlini, N.; Cubuk, E.~D.; Kurakin, A.; Sohn, K.; Zhang, H.; and
  Raffel, C. 2019{\natexlab{a}}.
\newblock ReMixMatch: Semi-Supervised Learning with Distribution Alignment and
  Augmentation Anchoring.
\newblock \emph{CoRR} abs/1911.09785.
\newblock \urlprefix\url{http://arxiv.org/abs/1911.09785}.

\bibitem[{Berthelot et~al.(2019{\natexlab{b}})Berthelot, Carlini, Goodfellow,
  Papernot, Oliver, and Raffel}]{2019MixMatch}
Berthelot, D.; Carlini, N.; Goodfellow, I.~J.; Papernot, N.; Oliver, A.; and
  Raffel, C. 2019{\natexlab{b}}.
\newblock MixMatch: {A} Holistic Approach to Semi-Supervised Learning.
\newblock In Wallach, H.~M.; Larochelle, H.; Beygelzimer, A.;
  d'Alch{\'{e}}{-}Buc, F.; Fox, E.~B.; and Garnett, R., eds., \emph{Advances in
  Neural Information Processing Systems 32: Annual Conference on Neural
  Information Processing Systems 2019, NeurIPS 2019, December 8-14, 2019,
  Vancouver, BC, Canada}, 5050--5060.
\newblock
  \urlprefix\url{https://proceedings.neurips.cc/paper/2019/hash/1cd138d0499a68f4bb72bee04bbec2d7-Abstract.html}.

\bibitem[{Bochkovskiy, Wang, and
  Liao(2020)}]{DBLP:journals/corr/abs-2004-10934}
Bochkovskiy, A.; Wang, C.; and Liao, H.~M. 2020.
\newblock YOLOv4: Optimal Speed and Accuracy of Object Detection.
\newblock \emph{CoRR} abs/2004.10934.
\newblock \urlprefix\url{https://arxiv.org/abs/2004.10934}.

\bibitem[{Cao et~al.(2020)Cao, Chen, Loy, and Lin}]{cao2019prime}
Cao, Y.; Chen, K.; Loy, C.~C.; and Lin, D. 2020.
\newblock Prime Sample Attention in Object Detection.
\newblock In \emph{2020 {IEEE/CVF} Conference on Computer Vision and Pattern
  Recognition, {CVPR} 2020, Seattle, WA, USA, June 13-19, 2020}, 11580--11588.
  {IEEE}.
\newblock \doi{10.1109/CVPR42600.2020.01160}.
\newblock \urlprefix\url{https://doi.org/10.1109/CVPR42600.2020.01160}.

\bibitem[{Chen et~al.(2019)Chen, Li, Lin, See, Wang, Duan, Chen, He, and
  Zou}]{chen2020accurate}
Chen, K.; Li, J.; Lin, W.; See, J.; Wang, J.; Duan, L.; Chen, Z.; He, C.; and
  Zou, J. 2019.
\newblock Towards Accurate One-Stage Object Detection With AP-Loss.
\newblock In \emph{{IEEE} Conference on Computer Vision and Pattern
  Recognition, {CVPR} 2019, Long Beach, CA, USA, June 16-20, 2019}, 5119--5127.
  Computer Vision Foundation / {IEEE}.
\newblock \doi{10.1109/CVPR.2019.00526}.
\newblock
  \urlprefix\url{http://openaccess.thecvf.com/content\_CVPR\_2019/html/Chen\_Towards\_Accurate\_One-Stage\_Object\_Detection\_With\_AP-Loss\_CVPR\_2019\_paper.html}.

\bibitem[{Chen et~al.(2020)Chen, Kornblith, Norouzi, and Hinton}]{2020A}
Chen, T.; Kornblith, S.; Norouzi, M.; and Hinton, G.~E. 2020.
\newblock A Simple Framework for Contrastive Learning of Visual
  Representations.
\newblock In \emph{Proceedings of the 37th International Conference on Machine
  Learning, {ICML} 2020, 13-18 July 2020, Virtual Event}, volume 119 of
  \emph{Proceedings of Machine Learning Research}, 1597--1607. {PMLR}.
\newblock \urlprefix\url{http://proceedings.mlr.press/v119/chen20j.html}.

\bibitem[{{Dollar} et~al.(2012){Dollar}, {Wojek}, {Schiele}, and
  {Perona}}]{5975165}
{Dollar}, P.; {Wojek}, C.; {Schiele}, B.; and {Perona}, P. 2012.
\newblock Pedestrian Detection: An Evaluation of the State of the Art.
\newblock \emph{IEEE Transactions on Pattern Analysis and Machine Intelligence}
  34(4): 743--761.
\newblock \doi{10.1109/TPAMI.2011.155}.

\bibitem[{Duan et~al.(2019)Duan, Bai, Xie, Qi, Huang, and Tian}]{0CenterNet}
Duan, K.; Bai, S.; Xie, L.; Qi, H.; Huang, Q.; and Tian, Q. 2019.
\newblock CenterNet: Keypoint Triplets for Object Detection.
\newblock In \emph{2019 {IEEE/CVF} International Conference on Computer Vision,
  {ICCV} 2019, Seoul, Korea (South), October 27 - November 2, 2019},
  6568--6577. {IEEE}.
\newblock \doi{10.1109/ICCV.2019.00667}.
\newblock \urlprefix\url{https://doi.org/10.1109/ICCV.2019.00667}.

\bibitem[{Everingham et~al.(2010)Everingham, Gool, Williams, Winn, and
  Zisserman}]{2010The}
Everingham, M.; Gool, L.~V.; Williams, C.; Winn, J.; and Zisserman, A. 2010.
\newblock The Pascal Visual Object Classes (VOC) Challenge.
\newblock \emph{International Journal of Computer Vision} 88(2): 303--338.

\bibitem[{Girshick(2015)}]{2015Fast}
Girshick, R.~B. 2015.
\newblock Fast {R-CNN}.
\newblock In \emph{2015 {IEEE} International Conference on Computer Vision,
  {ICCV} 2015, Santiago, Chile, December 7-13, 2015}, 1440--1448. {IEEE}
  Computer Society.
\newblock \doi{10.1109/ICCV.2015.169}.
\newblock \urlprefix\url{https://doi.org/10.1109/ICCV.2015.169}.

\bibitem[{Girshick et~al.(2014)Girshick, Donahue, Darrell, and
  Malik}]{2013Rich}
Girshick, R.~B.; Donahue, J.; Darrell, T.; and Malik, J. 2014.
\newblock Rich Feature Hierarchies for Accurate Object Detection and Semantic
  Segmentation.
\newblock In \emph{2014 {IEEE} Conference on Computer Vision and Pattern
  Recognition, {CVPR} 2014, Columbus, OH, USA, June 23-28, 2014}, 580--587.
  {IEEE} Computer Society.
\newblock \doi{10.1109/CVPR.2014.81}.
\newblock \urlprefix\url{https://doi.org/10.1109/CVPR.2014.81}.

\bibitem[{He et~al.(2017)He, Gkioxari, Doll{\'{a}}r, and Girshick}]{2017Mask}
He, K.; Gkioxari, G.; Doll{\'{a}}r, P.; and Girshick, R.~B. 2017.
\newblock Mask {R-CNN}.
\newblock In \emph{{IEEE} International Conference on Computer Vision, {ICCV}
  2017, Venice, Italy, October 22-29, 2017}, 2980--2988. {IEEE} Computer
  Society.
\newblock \doi{10.1109/ICCV.2017.322}.
\newblock \urlprefix\url{https://doi.org/10.1109/ICCV.2017.322}.

\bibitem[{Iscen et~al.(2019)Iscen, Tolias, Avrithis, and Chum}]{2019Label}
Iscen, A.; Tolias, G.; Avrithis, Y.; and Chum, O. 2019.
\newblock Label Propagation for Deep Semi-Supervised Learning.
\newblock In \emph{{IEEE} Conference on Computer Vision and Pattern
  Recognition, {CVPR} 2019, Long Beach, CA, USA, June 16-20, 2019}, 5070--5079.
  Computer Vision Foundation / {IEEE}.
\newblock \doi{10.1109/CVPR.2019.00521}.
\newblock
  \urlprefix\url{http://openaccess.thecvf.com/content\_CVPR\_2019/html/Iscen\_Label\_Propagation\_for\_Deep\_Semi-Supervised\_Learning\_CVPR\_2019\_paper.html}.

\bibitem[{Jeong et~al.(2019)Jeong, Lee, Kim, and Kwak}]{2019csd}
Jeong, J.; Lee, S.; Kim, J.; and Kwak, N. 2019.
\newblock Consistency-based Semi-supervised Learning for Object detection.
\newblock In Wallach, H.~M.; Larochelle, H.; Beygelzimer, A.;
  d'Alch{\'{e}}{-}Buc, F.; Fox, E.~B.; and Garnett, R., eds., \emph{Advances in
  Neural Information Processing Systems 32: Annual Conference on Neural
  Information Processing Systems 2019, NeurIPS 2019, December 8-14, 2019,
  Vancouver, BC, Canada}, 10758--10767.
\newblock
  \urlprefix\url{https://proceedings.neurips.cc/paper/2019/hash/d0f4dae80c3d0277922f8371d5827292-Abstract.html}.

\bibitem[{Jeong et~al.(2020)Jeong, Verma, Hyun, Kannala, and Kwak}]{2020isd}
Jeong, J.; Verma, V.; Hyun, M.; Kannala, J.; and Kwak, N. 2020.
\newblock Interpolation-based semi-supervised learning for object detection.
\newblock \emph{ArXiv} abs/2006.02158.

\bibitem[{Laine and Aila(2017)}]{laine2017temporal}
Laine, S.; and Aila, T. 2017.
\newblock Temporal Ensembling for Semi-Supervised Learning.
\newblock In \emph{5th International Conference on Learning Representations,
  {ICLR} 2017, Toulon, France, April 24-26, 2017, Conference Track
  Proceedings}. OpenReview.net.
\newblock \urlprefix\url{https://openreview.net/forum?id=BJ6oOfqge}.

\bibitem[{Law and Deng(2018)}]{DBLP:journals/corr/abs-1808-01244}
Law, H.; and Deng, J. 2018.
\newblock CornerNet: Detecting Objects as Paired Keypoints.
\newblock \emph{ArXiv} abs/1808.01244.

\bibitem[{Lin et~al.(2017)Lin, Goyal, Girshick, He, and
  Doll{\'{a}}r}]{2017Focal}
Lin, T.; Goyal, P.; Girshick, R.~B.; He, K.; and Doll{\'{a}}r, P. 2017.
\newblock Focal Loss for Dense Object Detection.
\newblock In \emph{{IEEE} International Conference on Computer Vision, {ICCV}
  2017, Venice, Italy, October 22-29, 2017}, 2999--3007. {IEEE} Computer
  Society.
\newblock \doi{10.1109/ICCV.2017.324}.
\newblock \urlprefix\url{https://doi.org/10.1109/ICCV.2017.324}.

\bibitem[{Lin et~al.(2014)Lin, Maire, Belongie, Hays, Perona, Ramanan,
  Doll{\'a}r, and Zitnick}]{2014Microsoft}
Lin, T.-Y.; Maire, M.; Belongie, S.; Hays, J.; Perona, P.; Ramanan, D.;
  Doll{\'a}r, P.; and Zitnick, C.~L. 2014.
\newblock Microsoft COCO: Common Objects in Context.
\newblock In Fleet, D.; Pajdla, T.; Schiele, B.; and Tuytelaars, T., eds.,
  \emph{Computer Vision -- ECCV 2014}, 740--755. Cham: Springer International
  Publishing.
\newblock ISBN 978-3-319-10602-1.

\bibitem[{Liu et~al.(2021)Liu, Ma, He, Kuo, Chen, Zhang, Wu, Kira, and
  Vajda}]{liu2021unbiased}
Liu, Y.; Ma, C.-Y.; He, Z.; Kuo, C.-W.; Chen, K.; Zhang, P.; Wu, B.; Kira, Z.;
  and Vajda, P. 2021.
\newblock Unbiased Teacher for Semi-Supervised Object Detection.
\newblock \emph{ArXiv} abs/2102.09480.

\bibitem[{Oksuz et~al.(2020)Oksuz, Cam, Kalkan, and Akbas}]{imbalance}
Oksuz, K.; Cam, B.~C.; Kalkan, S.; and Akbas, E. 2020.
\newblock {Imbalance Problems in Object Detection: A Review}.
\newblock \emph{Transactions on Pattern Analysis and Machine Intelligence
  (TPAMI)} 1--1.

\bibitem[{Ouyang et~al.(2016)Ouyang, Wang, Zhang, and Yang}]{2016Factors}
Ouyang, W.; Wang, X.; Zhang, C.; and Yang, X. 2016.
\newblock Factors in Finetuning Deep Model for Object Detection with Long-Tail
  Distribution.
\newblock In \emph{2016 {IEEE} Conference on Computer Vision and Pattern
  Recognition, {CVPR} 2016, Las Vegas, NV, USA, June 27-30, 2016}, 864--873.
  {IEEE} Computer Society.
\newblock \doi{10.1109/CVPR.2016.100}.
\newblock \urlprefix\url{https://doi.org/10.1109/CVPR.2016.100}.

\bibitem[{Pang et~al.(2019)Pang, Chen, Shi, Feng, Ouyang, and Lin}]{2020Libra}
Pang, J.; Chen, K.; Shi, J.; Feng, H.; Ouyang, W.; and Lin, D. 2019.
\newblock Libra {R-CNN:} Towards Balanced Learning for Object Detection.
\newblock In \emph{{IEEE} Conference on Computer Vision and Pattern
  Recognition, {CVPR} 2019, Long Beach, CA, USA, June 16-20, 2019}, 821--830.
  Computer Vision Foundation / {IEEE}.
\newblock \doi{10.1109/CVPR.2019.00091}.
\newblock
  \urlprefix\url{http://openaccess.thecvf.com/content\_CVPR\_2019/html/Pang\_Libra\_R-CNN\_Towards\_Balanced\_Learning\_for\_Object\_Detection\_CVPR\_2019\_paper.html}.

\bibitem[{Peng et~al.(2020)Peng, Bu, Sun, Zhang, Tan, and
  Yan}]{peng2020largescale}
Peng, J.; Bu, X.; Sun, M.; Zhang, Z.; Tan, T.; and Yan, J. 2020.
\newblock Large-Scale Object Detection in the Wild From Imbalanced
  Multi-Labels.
\newblock In \emph{2020 {IEEE/CVF} Conference on Computer Vision and Pattern
  Recognition, {CVPR} 2020, Seattle, WA, USA, June 13-19, 2020}, 9706--9715.
  {IEEE}.
\newblock \doi{10.1109/CVPR42600.2020.00973}.
\newblock \urlprefix\url{https://doi.org/10.1109/CVPR42600.2020.00973}.

\bibitem[{Redmon et~al.(2016)Redmon, Divvala, Girshick, and Farhadi}]{2016You}
Redmon, J.; Divvala, S.~K.; Girshick, R.~B.; and Farhadi, A. 2016.
\newblock You Only Look Once: Unified, Real-Time Object Detection.
\newblock In \emph{2016 {IEEE} Conference on Computer Vision and Pattern
  Recognition, {CVPR} 2016, Las Vegas, NV, USA, June 27-30, 2016}, 779--788.
  {IEEE} Computer Society.
\newblock \doi{10.1109/CVPR.2016.91}.
\newblock \urlprefix\url{https://doi.org/10.1109/CVPR.2016.91}.

\bibitem[{Ren et~al.(2015)Ren, He, Girshick, and Sun}]{2017Faster}
Ren, S.; He, K.; Girshick, R.~B.; and Sun, J. 2015.
\newblock Faster {R-CNN:} Towards Real-Time Object Detection with Region
  Proposal Networks.
\newblock In Cortes, C.; Lawrence, N.~D.; Lee, D.~D.; Sugiyama, M.; and
  Garnett, R., eds., \emph{Advances in Neural Information Processing Systems
  28: Annual Conference on Neural Information Processing Systems 2015, December
  7-12, 2015, Montreal, Quebec, Canada}, 91--99.
\newblock
  \urlprefix\url{https://proceedings.neurips.cc/paper/2015/hash/14bfa6bb14875e45bba028a21ed38046-Abstract.html}.

\bibitem[{{Russakovsky}, {Li}, and {Fei-Fei}(2015)}]{7298824}
{Russakovsky}, O.; {Li}, L.; and {Fei-Fei}, L. 2015.
\newblock Best of both worlds: Human-machine collaboration for object
  annotation.
\newblock In \emph{2015 IEEE Conference on Computer Vision and Pattern
  Recognition (CVPR)}, 2121--2131.
\newblock \doi{10.1109/CVPR.2015.7298824}.

\bibitem[{Sajjadi, Javanmardi, and Tasdizen(2016)}]{2016Regularization}
Sajjadi, M.; Javanmardi, M.; and Tasdizen, T. 2016.
\newblock Regularization With Stochastic Transformations and Perturbations for
  Deep Semi-Supervised Learning.
\newblock In Lee, D.~D.; Sugiyama, M.; von Luxburg, U.; Guyon, I.; and Garnett,
  R., eds., \emph{Advances in Neural Information Processing Systems 29: Annual
  Conference on Neural Information Processing Systems 2016, December 5-10,
  2016, Barcelona, Spain}, 1163--1171.
\newblock
  \urlprefix\url{https://proceedings.neurips.cc/paper/2016/hash/30ef30b64204a3088a26bc2e6ecf7602-Abstract.html}.

\bibitem[{Sohn et~al.(2020)Sohn, Zhang, Li, Zhang, Lee, and
  Pfister}]{sohn2020detection}
Sohn, K.; Zhang, Z.; Li, C.-L.; Zhang, H.; Lee, C.-Y.; and Pfister, T. 2020.
\newblock A Simple Semi-Supervised Learning Framework for Object Detection.
\newblock In \emph{arXiv:2005.04757}.

\bibitem[{Takeru et~al.(2018)Takeru, Shin-Ichi, Shin, and
  Masanori}]{2018Virtual}
Takeru, M.; Shin-Ichi, M.; Shin, I.; and Masanori, K. 2018.
\newblock Virtual Adversarial Training: A Regularization Method for Supervised
  and Semi-Supervised Learning.
\newblock \emph{IEEE Transactions on Pattern Analysis and Machine Intelligence}
  1--1.

\bibitem[{Tarvainen and Valpola(2017)}]{tarvainen2018mean}
Tarvainen, A.; and Valpola, H. 2017.
\newblock Mean teachers are better role models: Weight-averaged consistency
  targets improve semi-supervised deep learning results.
\newblock In Guyon, I.; von Luxburg, U.; Bengio, S.; Wallach, H.~M.; Fergus,
  R.; Vishwanathan, S. V.~N.; and Garnett, R., eds., \emph{Advances in Neural
  Information Processing Systems 30: Annual Conference on Neural Information
  Processing Systems 2017, December 4-9, 2017, Long Beach, CA, {USA}},
  1195--1204.
\newblock
  \urlprefix\url{https://proceedings.neurips.cc/paper/2017/hash/68053af2923e00204c3ca7c6a3150cf7-Abstract.html}.

\bibitem[{Xie et~al.(2020{\natexlab{a}})Xie, Dai, Hovy, Luong, and
  Le}]{xie2020unsupervised}
Xie, Q.; Dai, Z.; Hovy, E.~H.; Luong, T.; and Le, Q. 2020{\natexlab{a}}.
\newblock Unsupervised Data Augmentation for Consistency Training.
\newblock In Larochelle, H.; Ranzato, M.; Hadsell, R.; Balcan, M.; and Lin, H.,
  eds., \emph{Advances in Neural Information Processing Systems 33: Annual
  Conference on Neural Information Processing Systems 2020, NeurIPS 2020,
  December 6-12, 2020, virtual}.
\newblock
  \urlprefix\url{https://proceedings.neurips.cc/paper/2020/hash/44feb0096faa8326192570788b38c1d1-Abstract.html}.

\bibitem[{Xie et~al.(2020{\natexlab{b}})Xie, Luong, Hovy, and
  Le}]{xie2020selftraining}
Xie, Q.; Luong, M.; Hovy, E.~H.; and Le, Q.~V. 2020{\natexlab{b}}.
\newblock Self-Training With Noisy Student Improves ImageNet Classification.
\newblock In \emph{2020 {IEEE/CVF} Conference on Computer Vision and Pattern
  Recognition, {CVPR} 2020, Seattle, WA, USA, June 13-19, 2020}, 10684--10695.
  {IEEE}.
\newblock \doi{10.1109/CVPR42600.2020.01070}.
\newblock \urlprefix\url{https://doi.org/10.1109/CVPR42600.2020.01070}.

\bibitem[{Yun et~al.(2019)Yun, Han, Chun, Oh, Yoo, and Choe}]{0CutMix}
Yun, S.; Han, D.; Chun, S.; Oh, S.~J.; Yoo, Y.; and Choe, J. 2019.
\newblock CutMix: Regularization Strategy to Train Strong Classifiers With
  Localizable Features.
\newblock In \emph{2019 {IEEE/CVF} International Conference on Computer Vision,
  {ICCV} 2019, Seoul, Korea (South), October 27 - November 2, 2019},
  6022--6031. {IEEE}.
\newblock \doi{10.1109/ICCV.2019.00612}.
\newblock \urlprefix\url{https://doi.org/10.1109/ICCV.2019.00612}.

\bibitem[{Zhang et~al.(2018)Zhang, Ciss{\'{e}}, Dauphin, and
  Lopez{-}Paz}]{2017mixup}
Zhang, H.; Ciss{\'{e}}, M.; Dauphin, Y.~N.; and Lopez{-}Paz, D. 2018.
\newblock mixup: Beyond Empirical Risk Minimization.
\newblock In \emph{6th International Conference on Learning Representations,
  {ICLR} 2018, Vancouver, BC, Canada, April 30 - May 3, 2018, Conference Track
  Proceedings}. OpenReview.net.
\newblock \urlprefix\url{https://openreview.net/forum?id=r1Ddp1-Rb}.

\bibitem[{Zhou et~al.(2021)Zhou, Yu, Wang, Qian, and
  Li}]{zhou2021instantteaching}
Zhou, Q.; Yu, C.; Wang, Z.; Qian, Q.; and Li, H. 2021.
\newblock Instant-Teaching: An End-to-End Semi-Supervised Object Detection
  Framework.
\newblock \emph{CoRR} abs/2103.11402.
\newblock \urlprefix\url{https://arxiv.org/abs/2103.11402}.

\end{thebibliography}
\end{document}